\documentclass[12pt]{article}

\usepackage{amssymb,amsmath,amsfonts,eurosym,geometry,ulem,graphicx,caption,color,setspace,sectsty,comment,footmisc,caption,
pdflscape,subfigure,array,hyperref}
\usepackage[T1]{fontenc}
\usepackage[utf8]{inputenc}
\usepackage{lmodern}

\usepackage{url}            % simple URL typesetting
\usepackage{booktabs}       % professional-quality tables
\usepackage{amsfonts}       % blackboard math symbols
\usepackage{nicefrac}       % compact symbols for 1/2, etc.
\usepackage{microtype}      % microtypography
\usepackage{xcolor}         % colors
\usepackage{amsthm,mathtools}
\usepackage{graphicx}

\usepackage{makecell}
\usepackage{float}

\usepackage[english]{babel}
\usepackage{csquotes}

\usepackage[
backend=biber,
style=apa,
]{biblatex}

\newtheorem{theorem}{Theorem}
\newtheorem{corollary}[theorem]{Corollary}

\newtheorem{definition}{Definition}
\newtheorem{lemma}{Lemma}

\newcolumntype{L}[1]{>{\raggedright\let\newline\\arraybackslash\hspace{0pt}}m{#1}}
\newcolumntype{C}[1]{>{\centering\let\newline\\arraybackslash\hspace{0pt}}m{#1}}
\newcolumntype{R}[1]{>{\raggedleft\let\newline\\arraybackslash\hspace{0pt}}m{#1}}

\newcommand{\norm}[1]{\left\lVert#1\right\rVert}

\begin{document}

\begin{titlepage}
\title{Neural SDEs for Conditional Time Series Generation and the Signature-Wasserstein-1 metric}
\author{Pere Díaz Lozano \\\vspace{-0.5em}
    \small Departament de Matemàtiques \\\vspace{-0.5em}
   \small Universitat Autònoma de Barcelona \\\vspace{-0.5em}
  \small \texttt{pere.diaz@uab.cat} 
  \and
  Toni Lozano Bagén \\\vspace{-0.5em}
    \small Departament de Matemàtiques \\\vspace{-0.5em}
   \small Universitat Autònoma de Barcelona \\\vspace{-0.5em}
  \small \texttt{tonilb@mat.uab.cat}
  \and  Josep Vives \\\vspace{-0.5em}
    \small Departament de Matemàtiques i Informàtica \\\vspace{-0.5em}
   \small Universitat de Barcelona \\\vspace{-0.5em}
  \small \texttt{josep.vives@ub.edu}  }
\date{\today}
\maketitle
\begin{abstract}
  (Conditional) Generative Adversarial Networks (GANs) have found great success in recent years, due to their ability to approximate (conditional) distributions over extremely high dimensional spaces. However, they are highly unstable and computationally expensive to train, especially in the time series setting. Recently, it has been proposed the use of a key object in rough path theory, called the signature of a path, which is able to convert the min-max formulation given by the (conditional) GAN framework into a classical minimization problem. However, this method is extremely expensive in terms of memory cost, sometimes even becoming prohibitive. To overcome this, we propose the use of \textit{Conditional Neural Stochastic Differential Equations}, which have a constant memory cost as a function of depth, being more memory efficient than traditional deep learning architectures. We empirically test that this proposed model is more efficient than other classical approaches, both in terms of memory cost and computational time, and that it usually outperforms them in terms of performance. 
\end{abstract}
\\
\thispagestyle{empty}
\end{titlepage}
\pagebreak \newpage

\section{Introduction}

The simplest approach to perform time series forecasting is to only be interested in a deterministic response, which is usually thought of as the mean of possible outcomes. Consequently, the model is unable to report any inherent uncertainty, which is a major shortcoming for most real world applications.

Recent work has ought to solve this by turning their attention to Generative Adversarial Networks (GANs) (\cite{pmlr-v70-arjovsky17a}, \cite{GAN}). Their basic idea is to train two networks against each other: 
\begin{itemize}
    \item \textbf{The generator:} it generates samples from a distribution by sending random noise through a parameterized model.
    
    \item \textbf{The adversarial network:} its job is to approximate a loss function between the real data distribution and the distribution produced by the generator. 

\end{itemize}

The training algorithm consists in training the adversarial network for a certain numbers of steps, and once a good enough approximation of the loss function we are interested in is reached, then perform one step on the parameters of the generator. 

The original GAN formulation was set such that the adversarial network approximated the Jensen Shannon divergence \parencite{GAN}. However, in our case we will focus on the Wasserstein GAN formulation, where the adversarial network (called the \textit{critic}) aims at approximating the Wasserstein-1 distance \parencite{pmlr-v70-arjovsky17a}.

Although GANs have had great success in approximating probability measures over extremely high dimensional spaces, especially in Computer Vision, they are very unstable to train. Moreover, the fact that one needs to train an adversarial network before performing one step on the generator means that the training time is considerably large. 

\subsection{Signature-Wasserstein-1 metric}

A more efficient approach would be to approximate the loss function in a closed form way, without the need of an iterative method. In this direction, \cite{nihao} proposed the use of the \textit{signature transform}, a fundamental object from rough path theory which captures many of the most important analytic and geometric properties of a given path\footnote{Although in practice we are usually given a set of discrete streams of data, we can always embed them into the set of continuous paths by performing some kind of interpolation scheme. Therefore, we will not really differentiate between these two concepts.}. By using it, one is able to derive an expression that is able to approximate uniformly well the Wasserstein-1 distance between two distributions on path space in a closed form non-parametric way. In doing so, all of the drawbacks that are inherent to the (Wasserstein) GAN formulation are overcome, while still maintaining its strong theoretical guarantees. This new framework is called the \textit{Signature-Wasserstein GAN} (SigWGAN).

In \cite{CSWGAN}, the authors proposed the \textit{Conditional Signature-Wasserstein GAN} (SigCWGAN), which is a modification of the SigWGAN to learn instead conditional distributions, as the ones we are interested in when doing non-deterministic time series forecasting. However, since this algorithm needs to perform a Montecarlo procedure for each data sample, the memory cost is dramatically increased, sometimes even becoming prohibitive. They elude this by assuming that the distribution of the observed time series has an autoregressive structure, with the next observed value depending only on the previous $q>0$, with $q$ relatively small. 

\subsection{Contributions}

In order to overcome this increase in the memory cost, we propose the use of \textit{Neural Stochastic Differential Equations} (Neural SDEs) (see \cite{kidger2021neuralsde}, \cite{pmlr-v108-li20i}), which are essentially generative models formed by classical stochastic differential equations whose vector fields are parameterized by neural networks. The main advantage over traditional deep learning architectures is the fact that one is able to backpropagate through a Neural SDE without the need to store the intermediate quantities produced in the forward pass, being more memory efficient than traditional deep learning models. 

By encoding the path we want to condition on, and by making the initial condition of the Neural SDE depend on this codification, we introduce what we call \textit{Conditional Neural Stochastic Differential Equations} (CNSDEs), which produce conditional distributions in path space. 

When using all of the above, we are able to formulate a model and a training algorithm for approximating conditional distributions on path space, which is both mathematically well founded and practically more efficient. Furthermore, we will test it against other more traditional baselines, concluding that in most cases it outperforms them according to several metrics. 

\section{Background}

\subsection{The Signature-Wasserstein-1 metric}
Let $sig(x)$, $sig^{N}(x)$ denote the signature and the truncated signature of order $N$ of a continuous path $x$, respectively, both with the basepoint and time augmentations (see \cite{DBLP:journals/corr/abs-2006-00873}).

Let $\mu$, $\nu$ be two distributions in path space. The Kantorovich-Rubinstein duality states that the Wasserstein-1 distance can be calculated as
\begin{gather*}
    W_{1}(\mu, \nu) = \sup_{\norm{F}_{L} \leq 1} \Big\{ \mathbb{E}_{x \sim \mu}[ F(x)] - \mathbb{E}_{x \sim \nu}[ F(x)] \Big\}
\end{gather*}
where the supremum is over all the $1-$Lipschitz functions. 

In \cite{nihao}, \cite{CSWGAN} the authors use the signature transform and its universal non-linearity to derive a closed form expression that uniformly approximates the Wasserstein-1 distance,
\begin{gather}\label{closedform1}
    sig^{N}W_{1}(\mu, \nu) = \norm{ \mathbb{E}_{x \sim \mu}[ sig^{N}(x)] - \mathbb{E}_{x \sim \nu}[ sig^{N}(x)]}_{2}
\end{gather}
where $\norm{\cdot}_{2}$ denotes the $L^{2}$ norm. This approximation is called the truncated Signature-Wasserstein-1 metric of order $N$.

Sections \ref{signaturesection} and \ref{sigw1section} in the supplementary material provide an in depth derivation of (\ref{closedform1}).

\subsection{The Conditional Signature-Wasserstein GAN algorithm}

In the unconditional case, whenever we intend to compute (\ref{closedform1}) from a set of given data, we can simply approximate both expected signatures by their empirical average, by performing a Montecarlo simulation.

However, if instead we are interested in approximating conditional distributions, then one is not able to perform a Montecarlo simulation to approximate the expected signatures of the conditional distributions given by the data, since most of the times we are only handled one sample.

By using many of the properties of the signature, one is able to prove that the conditional expected truncated signature can be approximated arbitrarily well by applying a linear map on the truncated signature of the conditioning path $x$,
\begin{gather}\label{closedform2}
    \hat{\mathbb{E}}_{y \sim Y|x}[sig^{M}(y)] = \hat{\ell}(sig^{N}(x))
\end{gather}
which can be approximated from the data by standard linear regression techniques. An in depth derivation of (\ref{closedform2}) is given in Section \ref{cswgansection}.

The resulting training algorithm is called the Conditional Signature-Wasserstein GAN (SigCWGAN) (see \cite[Algorithm 2]{CSWGAN}).
\begin{figure}[h!]
    \centering
    \includegraphics[width=13cm]{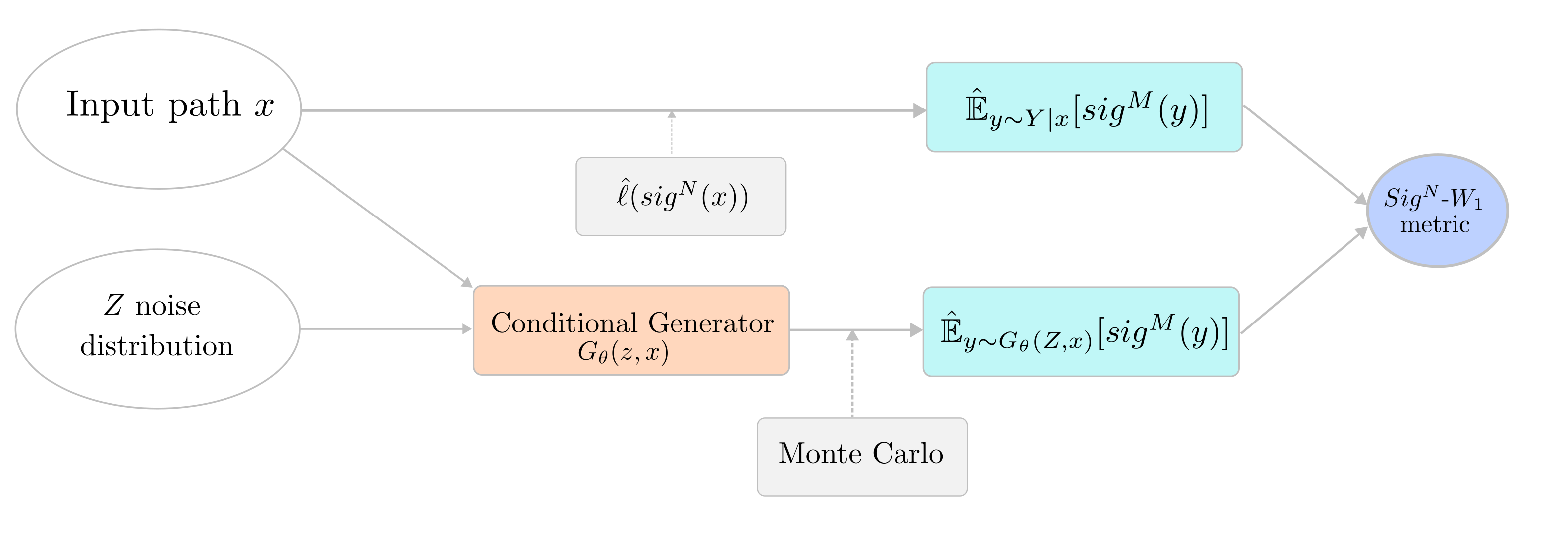}
    \caption{Flowchart of the SigCWGAN algorithm. }
\end{figure}
\subsubsection{Disadvantages of the SigCWGAN algorithm}

One of the main drawbacks of the SigCWGAN algorithm is the Montecarlo procedure needed to estimate the expected signature of the conditional generator for each sample in the minibatch, which considerably increases the memory cost. This is in contrast to the general conditional GAN setting, where both the generator and the critic simply take as input the corresponding sample $x$ \parencite{2014arXiv1411.1784M}. Therefore, the conditional Wasserstein GAN algorithm (CWGAN) has the same memory cost as the unconditional WGAN approach.

This means that the SigCWGAN algorithm consumes much more memory that the traditional CWGAN approach, and as a consequence sometimes we need to decrease the batch size or the complexity of the model. As we will see in the following pages, this can be remedied by using a Neural Stochastic Differential Equation as a generator, which are much more memory efficient to train than traditional neural networks architectures.

\subsection{Neural Stochastic Differential Equations}

 \textit{Neural Stochastic Differential Equations} (Neural SDE) are models that arise from the union of stochastic differential equations and neural networks, two of the biggest paradigms in mathematical modelling, resulting in generative models that produce distributions in path space.
 
The basic structure of a Neural SDE is 
\begin{flalign*}
    Z_{0} & \sim \xi_{\theta}(V) \\
    dZ_{t} &= f_{\theta}(t, Z_{t})dt + g_{\theta}(t, Z_{t}) \circ dW_{t} \\
    Y_{t} &= \alpha_{\theta}Z_{t}+\beta_{\theta}
\end{flalign*}
with $V \sim \mathcal{N}(0, I_{d_{v}})$ drawn from a $d_{v}-$dimensional standard normal distribution, $\xi_{\theta}$, $f_{\theta}$, $g_{\theta}$ being neural networks and $\alpha_{\theta}, \beta_{\theta}$ being matrices of learnable weights. $W_{t}$ denotes the Wiener process, and the $\circ$ indicates that the integration is in the Stratonovich sense. 

 $Z$ represents the hidden state of the model. If it was the output, then the resulting stochastic process would satisfy a Markov property, which does not need to be true in general. This is the reason for the final readout linearity.

As we already said, the solution to a stochastic differential equation is a distribution in path space. However, just like with ODEs, computing it in an analytical form is almost always impossible.  Nonetheless, one can sample from it. This is what a numerical SDE solver does, returning a sampled path evaluated at a set of discrete time locations. 

Thus a Neural SDE fits the GAN setting, since evaluating its density is not possible and it is able to generate samples by sending random noise (in the form of the initial condition and the Wiener process) through a parameterized model. 

\subsubsection{Backpropagation through a Neural SDE}

If we intend to fit a Neural SDE to some data, we need an algorithm to compute gradients with respect to its parameters. We will briefly summarize two ways to do this:

\textbf{Discretize-then-optimize:} The most straightforward way consists in performing the usual backpropagation, differentiating through the internal operations of the differential equation solver, and therefore needing to store them in the forward pass. In the literature this is commonly referred to as \textit{discretize-then-optimize}, because we are directly optimizing the discretization we are using in practice. 

If we denote by $H$ the memory cost of recording the operations of one solver step, and by $N$ the number of steps, then the discretize-then-optimize method consumes about $\mathcal{O}(HN)$ memory. Notice that this is the memory cost of traditional deep learning models, such as Recurrent Neural Networks. 

\textbf{Reversible solvers:} Alternatively, if we use a reversible solver, we dramatically reduce the memory cost. Consider a differential equation solver, which in the forward pass iteratively computes the next step from the previous one, 
\begin{gather}
    (z_{t_{i}}, \alpha_{t_{i}}) \mapsto (z_{t_{i+1}}, \alpha_{t_{i+1}}),
\end{gather}
where $\{z_{t_{i}}\}_{i=1}^{n}$ is the numerical approximation of the solution to some differential equation, and $\alpha_{t_{i}}$ represents the intermediate quantities that the solver needs to store at step $i$ in order to compute the solution at the next step $i+1$. 

Then, a solver is said to be \textit{algebraically reversible} if it can compute the previous step from the next step,
\begin{gather}
    (z_{t_{i+1}}, \alpha_{t_{i+1}}) \mapsto (z_{t_{i}}, \alpha_{t_{i}}),
\end{gather}
by using a closed-form expression. Notice how, in this case, the intermediate values $(z_{t_{i}}, \alpha_{t_{i}})$ do not need to be stored in memory, and they can be recomputed in the backward pass. In this case, the memory cost is only $\mathcal{O}(H)$.

The full algorithm, along with the only known such SDE solver (called the reversible Heun method), can be found in \cite{NEURIPS2021_9ba196c7}.

 \section{Conditional Neural Stochastic Differential Equations}
 
In our case we are interested in generating distributions that are conditioned on some input path $x$. We propose to condition a Neural SDE by making the initial condition of the SDE depend on the path we want to condition on, following an encoder-decoder structure.
 
 \begin{definition}[Conditional Neural Stochastic Differential Equation]\label{cnsde}
Let 
\begin{flalign*}
    \xi_{\theta} &: \mathbb{R}^{d_{h}} \to \mathbb{R}^{d_{z}} \\
    f_{\theta} &: [0,T] \times \mathbb{R}^{d_{z}} \to \mathbb{R}^{d_{z}} \\
    g_{\theta} &: [0,T] \times \mathbb{R}^{d_{z}} \to \mathbb{R}^{d_{z} \times d_{w}} 
\end{flalign*}
be feedforward neural networks. Let $\phi: \mathcal{TS}([0,\tau]; \mathbb{R}^{d_{x}}) \to \mathbb{R}^{d_{h}}$ be a continuous map (with or without learnable parameters), with $\mathcal{TS}([0,\tau]; \mathbb{R}^{d_{x}})$ being the space of time series with timestamps in $[0,\tau]$ and $d_{x}-$dimensional observations. Then we define a Conditional Neural Stochastic Differential Equation (CNSDE) as 
\begin{flalign*}
    h &= \phi(x) \\
    Z_{0} & = \xi_{\theta}(h) \\
    dZ_{t} &= f_{\theta}(t, Z_{t})dt + g_{\theta}(t, Z_{t}) \circ dW_{t} \\
    Y_{t} &= \alpha_{\theta}Z_{t}+\beta_{\theta}
\end{flalign*}
where $\alpha_{\theta} \in \mathbb{R}^{d_{y}\times d_{z}}$ and $\beta_{\theta} \in \mathbb{R}^{d_{y}}$ are matrices of learnable weights.
\end{definition}

The following is an overview of a CNSDE. An input path $x$ (red) is fed into the model, which determines the initial condition of the SDE $z_{0}$. Then a trajectory $w$ is sampled from the Wiener process (blue). All of this is fed to the differential equation solver, which gives us the solution $z$ (green). After that we apply a final readout linearity, which produces the output of the model $y$ (purple). As a summary, $y$ is a sample from a distribution in path space that is conditioned on a given sample $x$. Whenever we change $x$, this distribution changes.

\begin{figure}[H]
    \centering
    \includegraphics[width=10cm]{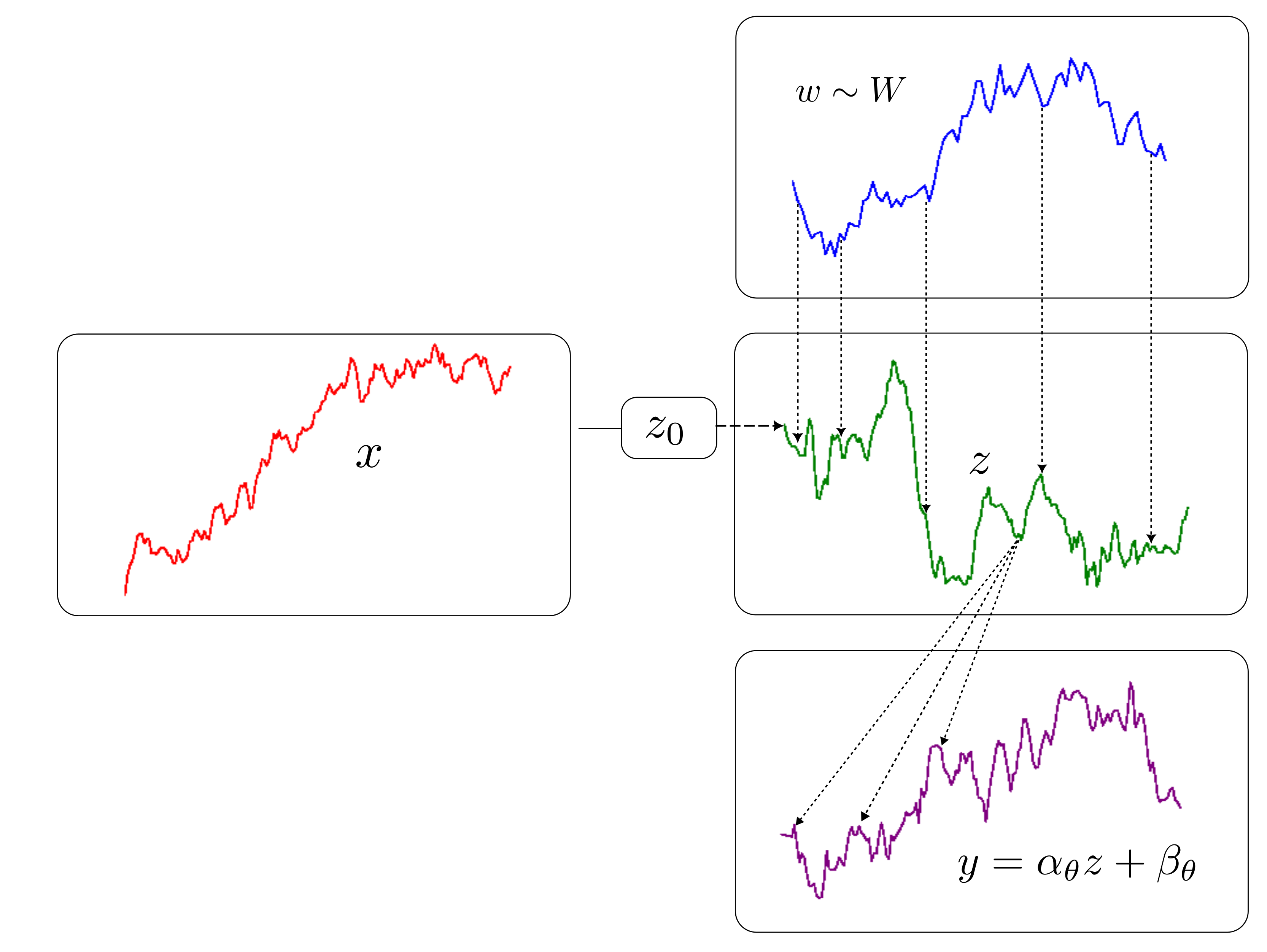}
    \caption{Overview of a Conditional Neural SDE. }
\end{figure}

Notice how the initial condition of the SDE is completely determined (once the parameters are fixed) by the input path $x$. An alternative approach is to add a random component to the initial condition $Z_{0}$, which usually enforces diversity and improves learning. This can be simply done by setting the initial condition to be 
\begin{flalign}
    h &= \phi(x) \\
    Z_{0} & \sim [\xi_{\theta}^{1}(h), \xi_{\theta}^{2}(V)]
\end{flalign}
with $\xi_{\theta}^{1} : \mathbb{R}^{d_{h}} \to \mathbb{R}^{d_{z}-k}$ and $\xi_{\theta}^{2} : \mathbb{R}^{d_{v}} \to \mathbb{R}^{k}$ being neural networks and $V \sim \mathcal{N}(0, I_{d_{v}})$. The larger the size of $k$ with respect to $d_{z}$, the more we will be enforcing diversity.

\section{Empirical Analysis}\label{empirical analysis}

The goal of this section is to compare the performance of the SigCWGAN algorithm and the CNSDE against some more traditional approaches which will serve as a baseline. All the code is available in the GitHub repository \href{https://github.com/pere98diaz/Neural-SDEs-for-Conditional-Time-Series-Generation-and-the-Signature-Wasserstein-1-metric}{https://github.com/pere98diaz/Neural-SDEs-for-Conditional-Time-Series-Generation-and-the-Signature-Wasserstein-1-metric}. 

We will compare the performance of three models:

\textbf{LSTM Conditional Wasserstein GAN:} the model that will serve as a pure baseline is based on \textit{Long short-term memory} (LSTM) networks \parencite{10.1162/neco.1997.9.8.1735}. It is trained by using the Conditional Wasserstein GAN algorithm, with the critic being also formed by LSTMs.

\textbf{LSTM Conditional Signature-Wasserstein GAN:} the second model is formed by the same conditional generator as the LSTM CWGAN model, but using the truncated Signature-Wasserstein-1 metric to approximate the Wasserstein-1 distance. To train the model we use the SigCWGAN algorithm.

\textbf{Neural SDE Conditional Signature-Wasserstein GAN:} the third model is formed by a Conditional Neural Stochastic Differential Equation as the generator, and uses the truncated Signature-Wasserstein-1 metric to approximate the Wasserstein-1 distance. We also use the SigCWGAN algorithm for training.

\subsection{Resources cost}\label{resources cost}

In this section we will be comparing the three models defined earlier in terms of two key resources: maximum memory allocated on the GPU (left hand side) and computational time required to perform one step on the generator (right hand side). The architectures of the three models are chosen so that they have a similar number of parameters.

\begin{figure}[H]
    \centering
    \includegraphics[width=13cm]{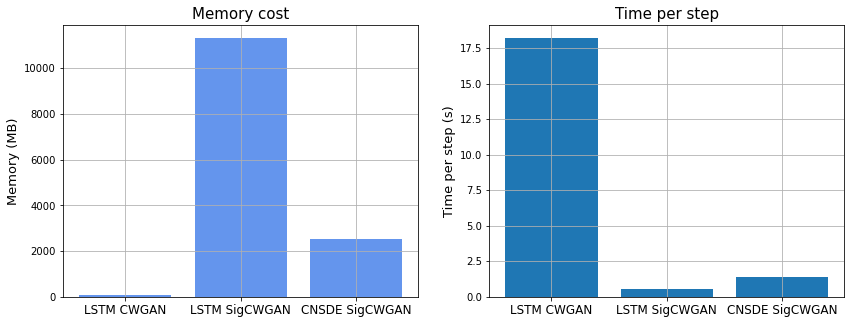}
    \caption{Memory consumption (MB) and time per step (s) for the three models.}
\end{figure}

We can see how, in terms of memory cost, the LSTM trained with the SigCWGAN algorithm is by far the most expensive one. This is mainly because of the Montecarlo procedure we need to perform for each sample on the minibatch. However, notice that when using the CNSDE as the conditional generator, the memory drops considerably, due to the use of a reversible solver to perform backpropagation. Unsurprisingly, the LSTM trained with the CWGAN algorithm is the cheapest one.

In terms of time per generator step, the most expensive model is the LSTM trained with the CWGAN algorithm. This is mainly because of the fact that we need to perform $k$ steps on the parameters of the critic before performing one step on the parameters of the generator (in our case, we picked the standard default value, which is $k=10$). The models trained with the SigCWGAN algorithm approximate the Wasserstein-1 loss in a closed form non-parametric way, which is the reason why it takes much less time to perform one generator step.

In conclusion, the most balanced model out of the three is the CNSDE trained with the SigCWGAN algorithm.

\subsection{Experiments}\label{experiments}

We performed experiments on four one dimensional different datasets, which will be described next. Each of the three models were trained three times, and we measured their performance in terms of the following statistics, all evaluated in an out-of-time test set.
\vspace{0.3cm}

\textbf{Classification error:} We train an LSTM whose task is to, given a pair of input/output streams, classify whether the output stream is real or generated. The architecture of this classifier is the same for each model. The worse the performance of this classifier, the better the generative model is. 
\vspace{0.3cm}

\textbf{Higher order Signature-Wasserstein-1 metric:} Similarly to the SigCWGAN algorithm, we compute the $L^{2}$-distance between the predicted and generated expected signature. However, in this case we truncate the signatures at higher levels than the ones we used during training: depth 6 for the input paths and depth 5 for the output paths. 
\vspace{0.3cm}

\textbf{Unordered Wasserstein-1 metric:} We compare the Wasserstein-1 distance between the real one dimensional distributions that are given by: \textbf{a)} taking the data points $y_{t}$ from the output streams, without considering that they are ordered in time. \textbf{b)} taking the differences between the data points in the output streams and the last value from the input stream, $y_{t}-x_{T}$, also without considering that they are ordered in time. \textbf{c)} For each stream, taking the largest difference given by the procedure in b). \textbf{d)} For each stream, taking the smallest difference given by the procedure in b).

\vspace{0.3cm}

\textbf{Extreme values metric:} Consider the empirical distribution given by the procedure we detailed in the Unordered Wasserstein-1 metric \textbf{c)}. Let $q$ indicate the value of a very high percentile of it, for example $95\%$. Then we estimate the probability, conditioned on the input path $x$, of producing a path $y$ such that $\max_{t} y_{t}-x_{T}$
is equal or over $q$. In some cases we will instead consider percentage increases, as $\max_{t} (y_{t}-x_{T})/x_{T}$. 

Just like we defined a way of measuring how good the model can detect a high possibility of an extremely large increment, we can do the same for an extremely value decrease, by considering $\min_{t} y_{t}-x_{T}$ instead (or $\min_{t} (y_{t}-x_{T})/x_{T}$) and setting the percentile to be very low, like $5\%$.
\vspace{0.3cm}

We should remark that evaluating the performance of a conditional generative model, especially in the time series framework, is very challenging, and none of the above statistics should be considered as ground truth metrics. Moreover, most of the time the metric we are interested in depends on the problem we have at hand, and the purpose and use we want to give to that model.

\subsubsection{AR(5) data}

The first dataset that we used was simulated from an autoregressive model of order $p=5$. We considered the problem of predicting the next 40 steps given the previous 60. The training time was set to a maximum of 2 hours. However, for the models using the SigCWGAN algorithm one is able to define an early stopping criteria, which was set to 1,000 steps without improvement. Moreover, at the end of training one can just keep the parameters that gave the best loss on a validation set. 

The next table indicates some general information on training, like the maximum memory allocated, the training time or the number of steps that were performed on the generator. The last column indicates the objective function that was used for training in the SigCWGAN models evaluated in the test set. For completeness, we also show it for the LSTM CWGAN, although it was not set to explicitly optimize it at all.

\begin{table}[H]\setlength\belowcaptionskip{-10pt}
\centering
\begin{tabular}{l c c c c}
\hline
             & Memory (MB) & Time (h)  & Steps   & Sig-$W_{1}$ loss   \\ \Xhline{2\arrayrulewidth}
LSTM CWGAN    &  $1945$     & $2.000 \pm 0.000$  & $566 \pm 15$  & $5.578 \pm 2.297$    \\
LSTM SigCWGAN &  $9931$     & $1.712 \pm 0.499$  & $8789 \pm 949$  & $2.804\pm 0.048$   \\ 
NSDE SigCWGAN &  $3764$     & $1.555 \pm 0.386$  &  $16839 \pm 1673$  &  $1.855 \pm 0.095$      \\ \Xhline{2\arrayrulewidth}
\end{tabular}
\caption{Some general information on the training process, AR(5) dataset.}\label{table 5.1}
\end{table}

Next we show the Area Under the Curve (AUC) and the accuracy obtained from training an LSTM to tell apart real from fake data. We can clearly see how the LSTM CWGAN outperforms the rest. This is  not a surprise at all, since in the GAN framework the generator is directly competing against another network whose job is precisely to distinguish real data from generated data.

The best model according to each metric is always marked as bold. 

\begin{table}[H]\setlength\belowcaptionskip{-10pt}
\begin{center}
\begin{tabular}{l c c}
\hline
             &  AUC           & Accuracy    \\ \Xhline{2\arrayrulewidth}
LSTM CWGAN    &    $\mathbf{0.747 \pm 0.144}$  & $\mathbf{0.685 \pm 0.121}$   \\
LSTM SigCWGAN & $0.866 \pm 0.053$  &  $0.775 \pm 0.043$  \\ 
NSDE SigCWGAN &   $0.959 \pm 0.021$ &    $0.873 \pm 0.031$ \\ \Xhline{2\arrayrulewidth}
\end{tabular}
\end{center}
\caption{Classification error, AR(5) dataset. }\label{table 5.2}
\end{table}

\begin{table}[H]\setlength\belowcaptionskip{-10pt}
\begin{center}
\begin{tabular}{l c c c}
\hline
             & HO Sig-$W_{1}$ metric &  $\text{EV}_{+}$ AUC & $\text{EV}_{-}$ AUC \\ \Xhline{2\arrayrulewidth}
LSTM CWGAN    &  $5.412 \pm 0.361$ &  $0.831 \pm 0.098$  & $0.834 \pm 0.130$           \\
LSTM SigCWGAN &  $5.092 \pm 0.005$  &  $0.958 \pm 0.039$ & $0.960 \pm 0.035$         \\ 
NSDE SigCWGAN &  $\mathbf{1.074 \pm 0.083}$   &  $\mathbf{0.988 \pm 0.004}$ & $\mathbf{0.982 \pm 0.013}$                         \\ \Xhline{2\arrayrulewidth}
\end{tabular}
\end{center}
\caption{The Higher order Signature-Wasserstein-1 metric and both the Extreme values metrics, AR(5) dataset. The selected percentiles for the $\text{EV}_{+}$ and $\text{EV}_{-}$ were 99\% and 1\%, respectively.}
\end{table}

\begin{table}[H]\setlength\belowcaptionskip{-10pt}
\begin{center}
\begin{tabular}{l c c c c}
\hline
             & $W_{1}$-a & $W_{1}$-b & $W_{1}$-c & $W_{1}$-d       \\ \Xhline{2\arrayrulewidth}
LSTM CWGAN    & $0.092 \pm 0.127$  & $0.061 \pm 0.075$  & $0.216 \pm 0.296$ & $0.236 \pm 0.301$         \\
LSTM SigCWGAN & $0.086 \pm 0.016$      & $0.056 \pm 0.004$  & $\mathbf{0.155 \pm 0.089}$  & $\mathbf{0.135 \pm 0.059}$            \\ 
NSDE SigCWGAN & $\mathbf{0.036 \pm 0.015}$      & $\mathbf{0.017 \pm 0.005}$   & $0.156 \pm 0.035$  & $0.165 \pm 0.061$             \\ \Xhline{2\arrayrulewidth}
\end{tabular}
\end{center}
\caption{The four unordered Wasserstein-1 metrics, AR(5) dataset.}
\end{table}

In conclusion, we see in Table \ref{table 5.1} that the Neural SDE did a better job minimizing the Sig-$W_{1}$ loss function than both LSTMs. This did not translate into a better performance in terms of the Classification error in Table \ref{table 5.2}, meaning that probably the Signature-Wasserstein-1 metric failed to produce a good enough approximation of the codification of the conditioned stochastic process. However, in terms of the rest of the metrics, the Neural SDE outperforms the other models, implying that it was able to encode many of its most important geometric properties.

\subsubsection{Seattle weather}
The second dataset was formed by daily observations of the maximum temperature reached in Seattle from 1948 to 2017\footnote{The dataset can be found in \href{https://www.kaggle.com/datasets/rtatman/did-it-rain-in-seattle-19482017}{https://www.kaggle.com/datasets/rtatman/did-it-rain-in-seattle-19482017}}. Our task will be to, given the last 60 days, predict the next 30. 

The training time was set to a maximum of 3 hours. For the models using the SigCWGAN we set an early stopping criteria of 1,000 steps without improvement. 

\begin{table}[H]\setlength\belowcaptionskip{-10pt}
\begin{center}
\begin{tabular}{l c c c c}
\hline
             & Memory (MB) & Time (h)  & Steps           & Sig-$W_{1}$ loss    \\ \Xhline{2\arrayrulewidth}
LSTM CWGAN    &  $2055$     & $3.000 \pm 0.000$        & $965 \pm 6$  & $6.944 \pm 0.360$   \\
LSTM SigCWGAN &  $8881$     & $0.937 \pm 0.201$  & $7501 \pm 1639$  & $3.119 \pm 1.009$     \\ 
NSDE SigCWGAN &  $4696$     & $1.317 \pm 0.460$  &   $4834 \pm 1626$      & $1.410 \pm 0.067$     \\ \Xhline{2\arrayrulewidth}
\end{tabular}
\end{center}
\caption{Some general information on the training procedure,
Seattle Weather dataset.}\label{table 5.5}
\end{table}

\newpage

\begin{table}[H]\setlength\belowcaptionskip{-10pt}
\begin{center}
\begin{tabular}{l c c}
\hline
             &  AUC           & Accuracy    \\ \Xhline{2\arrayrulewidth}
LSTM CWGAN    &  $\mathbf{0.617 \pm 0.060}$    & $\mathbf{0.573 \pm 0.033}$   \\
LSTM SigCWGAN & $0.715 \pm 0.114$  &  $0.652 \pm 0.087$  \\ 
NSDE SigCWGAN &   $0.732 \pm 0.038$ &    $0.660 \pm 0.031$ \\ \Xhline{2\arrayrulewidth}
\end{tabular}
\end{center}
\caption{Classification error, Seattle Weather dataset.}\label{table 5.6}
\end{table}

\begin{table}[H]\setlength\belowcaptionskip{-10pt}
\begin{center}
\begin{tabular}{l c c c}
\hline
             & HO Sig-$W_{1}$ &  $\text{EV}_{+}$ AUC & $\text{EV}_{-}$ AUC \\ \Xhline{2\arrayrulewidth}
LSTM CWGAN    &  $6.014 \pm 0.026$ &  $0.514 \pm 0.025$  & $0.577 \pm 0.047$           \\
LSTM SigCWGAN &  $5.465 \pm 0.265$  &  $0.808 \pm 0.099$ & $0.888 \pm 0.046$         \\ 
NSDE SigCWGAN &  $\mathbf{1.032 \pm 0.082}$   &  $\mathbf{0.905 \pm 0.001}$ & $\mathbf{0.940 \pm 0.003}$                         \\ \Xhline{2\arrayrulewidth}
\end{tabular}
\end{center}
\caption{The Signature-Wasserstein-1 metric and both the Extreme values metrics, Seattle Weather dataset. The selected percentiles for the $\text{EV}_{+}$ and $\text{EV}_{-}$ were 95\% and 5\%, respectively.}\label{table 5.7}
\end{table}

\begin{table}[H]\setlength\belowcaptionskip{-10pt}
\begin{center}
\begin{tabular}{l c c c c}
\hline
             & $W_{1}$-a & $W_{1}$-b & $W_{1}$-c & $W_{1}$-d      \\ \Xhline{2\arrayrulewidth}
LSTM CWGAN    & $0.061 \pm 0.007$  & $0.155 \pm 0.018$  & $0.251 \pm 0.018$ & $0.175 \pm 0.018$         \\
LSTM SigCWGAN & $0.060 \pm 0.011$      & $0.035 \pm 0.006$  & $0.094 \pm 0.030$  & $0.144 \pm 0.072$            \\ 
NSDE SigCWGAN & $\mathbf{0.047 \pm 0.002}$      & $\mathbf{0.028 \pm 0.001}$   & $\mathbf{0.062 \pm 0.012}$  & $\mathbf{0.055 \pm 0.008}$             \\ \Xhline{2\arrayrulewidth}
\end{tabular}
\end{center}
\caption{The four unordered Wasserstein-1 metrics, Seattle Weather dataset.}
\end{table}

The conclusions are pretty much similar to the ones we obtained for the AR(5) dataset. In Table \ref{table 5.5} we see that the Neural SDE did a way better job at minimizing the CSig-W loss function than the LSTMs. Since this did not translate to the Classification error performance showed in Table \ref{table 5.6}, this means that the Signature-Wasserstein-1 metric did not do a great job at codifying the conditional stochastic process. However, since the Neural SDE outperforms the other models in terms of the rest of the metrics, we also conclude that it succeeded in capturing many important geometric properties. 

We especially highlight the results showed in Table \ref{table 5.7}, where we can see how the LSTM trained with the WGAN algorithm completely failed to perform well in terms of detecting extreme values. This is in contrast to the models trained with the SigCWGAN method, which did a very good job. 

\subsubsection{Forex}
The third dataset was a Forex time series formed by observations of the bid price between the Euro and the Dollar. More specifically, at each timestamp it indicates the highest price a buyer will pay, in Dollars, to buy one Euro. 

The observations are spanned over weeks 19 and 20 of 2020, and are irregularly spaced. The NSDE SigCWGAN model is able to work with irregularly sampled data, since both the conditioner and the loss function are based on the signature transform. However, this is not the case for the other two models. In order to be able to compare them, we aggregated the data by computing the mean in each 30 seconds interval.

 Our task will be to, given the last 80 observations, predict the next 80. The training time was set to a maximum of 4 hours. For the models using the SigCWGAN algorithm, we set an early stopping criteria of 1,000 steps without improvement. 

\begin{table}[H]\setlength\belowcaptionskip{-10pt}
\begin{center}
\begin{tabular}{l c c c c}
\hline
             & Memory (MB) & Time (h)  & Steps           & Sig-$W_{1}$ loss    \\ \Xhline{2\arrayrulewidth}
LSTM CWGAN    &  $5818$     & $4.000 \pm 0.000$        & $718 \pm 6$  & $4.860 \pm 1.947$   \\
LSTM SigCWGAN &  $9209$     & $1.555 \pm 0.252$  & $10417 \pm 1664$  & $2.491 \pm 0.002$     \\ 
NSDE SigCWGAN &  $8697$     & $1.653 \pm 0.236$  &   $3001 \pm 433$      & $2.049 \pm 0.019$     \\ \Xhline{2\arrayrulewidth}
\end{tabular}
\end{center}
\caption{Some general information on the training procedure,
Forex dataset.}
\end{table}

\begin{table}[H]\setlength\belowcaptionskip{-10pt}
\begin{center}
\begin{tabular}{l c c}
\hline
             &  AUC           & Accuracy    \\ \Xhline{2\arrayrulewidth}
LSTM CWGAN    &  $0.962 \pm 0.087$    & $0.932 \pm 0.097$   \\
LSTM SigCWGAN & $0.838 \pm 0.104$  &  $0.768 \pm 0.118 $  \\ 
NSDE SigCWGAN &   $\mathbf{0.630 \pm 0.093}$ &    $\mathbf{0.587 \pm 0.060}$ \\ \Xhline{2\arrayrulewidth}
\end{tabular}
\end{center}
\caption{Classification error, Forex dataset.}
\end{table}

\begin{table}[H]\setlength\belowcaptionskip{-10pt}
\begin{center}
\begin{tabular}{l c c c}
\hline
             & HO Sig-$W_{1}$ &  $\text{EV}_{+}$ AUC & $\text{EV}_{-}$ AUC \\ \Xhline{2\arrayrulewidth}
LSTM CWGAN    &  $4.614 \pm 1.346$ &  $0.500 \pm 0.004$  & $0.510 \pm 0.017$           \\
LSTM SigCWGAN &  $3.010 \pm 0.004$  & $0.488 \pm 0.011$  & $0.499 \pm 0.005$        \\ 
NSDE SigCWGAN &  $\mathbf{2.704 \pm 0.077}$   &  $\mathbf{0.596 \pm 0.024}$ & $\mathbf{0.544 \pm 0.013}$                         \\ \Xhline{2\arrayrulewidth}
\end{tabular}
\end{center}
\caption{The Signature-Wasserstein-1 metric and both the Extreme values metrics, Forex dataset. The selected percentiles for the $\text{EV}_{+}$ and $\text{EV}_{-}$ were 90\% and 10\%, respectively.}\label{table 5.11}
\end{table}

\begin{table}[H]\setlength\belowcaptionskip{-10pt}
\begin{center}
\begin{tabular}{l c c c c}
\hline
             & $W_{1}$-a & $W_{1}$-b & $W_{1}$-c & $W_{1}$-d      \\ \Xhline{2\arrayrulewidth}
LSTM CWGAN    & $0.103 \pm 0.048$  & $0.107 \pm 0.046$  & $0.053 \pm 0.014$ & $0.146 \pm 0.052$         \\
LSTM SigCWGAN &  $\mathbf{0.012 \pm 0.001}$     & $0.020 \pm 0.003$   & $0.029 \pm 0.003$  & $0.037 \pm 0.007$          \\ 
NSDE SigCWGAN & $0.013 \pm 0.000$      & $\mathbf{0.019 \pm 0.002}$   & $\mathbf{0.028 \pm 0.003}$  & $\mathbf{0.015 \pm 0.002}$             \\ \Xhline{2\arrayrulewidth}
\end{tabular}
\end{center}
\caption{The four unordered Wasserstein-1 metrics, Forex dataset.}
\end{table}
Notice how, in contrast to the last two experiments, the NSDE SigCWGAN model clearly outperformed the rest of the models in terms of Classification error. We theorize that this is due to the increment, in terms of length, of both the input and output streams. However, further experiments should be conducted to test whether this is indeed the true reason or not.

In this problem we perhaps could be especially interested in knowing whether, given a known path $x$, there will be a large increment or reduction in a fixed time period. This is tested with the Extreme values metric, and the results of each model are shown in Table \ref{table 5.11}. We can see how the NSDE based model also outperforms the rest in terms of these metrics.

\subsubsection{IBEX35}
The final dataset was formed by the daily return (between 1993 and 2022) of the IBEX 35 (IBerian IndEX), which is the benchmark stock market index of the Bolsa de Madrid, Spain's principal stock exchange. 

 Our task will be to, given the last 30 observations, predict the next 15. The training time was set to a maximum of 2 hours. For the models using the SigCWGAN algorithm, we set an early stopping criteria of 1,000 steps without improvement. 

\begin{table}[H]\setlength\belowcaptionskip{-10pt}
\begin{center}
\begin{tabular}{l c c c c}
\hline
              & Memory (MB) & Time (h)           & Steps           & Sig-$W_{1}$ loss          \\ \Xhline{2\arrayrulewidth}
LSTM CWGAN    & $1786$      & $2 \pm 0.000$      & $1283 \pm 3$    & $2.837 \pm 0.297$     \\
LSTM SigCWGAN &  $9201$     & $1.442 \pm 0.295$  & $1334 \pm 2919$ & $1.788 \pm 0.020$     \\ 
NSDE SigCWGAN &  $5825$     & $1.647 \pm 0.423$  & $6917 \pm 1773$ & $1.811 \pm 0.016$     \\ \Xhline{2\arrayrulewidth}
\end{tabular}
\end{center}
\caption{Some general information on the training procedure,
IBEX35 dataset.}
\end{table}

\begin{table}[H]\setlength\belowcaptionskip{-10pt}
\begin{center}
\begin{tabular}{l c c}
\hline
              &  AUC                       & Accuracy                   \\ \Xhline{2\arrayrulewidth}
LSTM CWGAN    & $0.844 \pm 0.115$          & $0.765 \pm 0.104$          \\
LSTM SigCWGAN & $0.661 \pm 0.066$          & $0.618 \pm 0.052$          \\ 
NSDE SigCWGAN & $\mathbf{0.588 \pm 0.074}$ & $\mathbf{0.562 \pm 0.058}$ \\ \Xhline{2\arrayrulewidth}
\end{tabular}
\end{center}
\caption{Classification error, IBEX35 dataset.}
\end{table}

\begin{table}[H]\setlength\belowcaptionskip{-10pt}
\begin{center}
\begin{tabular}{l c c c}
\hline
              & HO Sig-$W_{1}$ metric                &  $\text{EV}_{+}$ AUC        & $\text{EV}_{-}$ AUC   \\ \Xhline{2\arrayrulewidth}
LSTM CWGAN    & $4.910 \pm 0.231$            &  $0.776 \pm 0.041$          & $0.507 \pm 0.118$           \\
LSTM SigCWGAN & $4.606 \pm 0.043$            &  $0.867 \pm 0.042$          & $0.642 \pm 0.048$           \\ 
NSDE SigCWGAN & $\mathbf{1.301 \pm 0.367}$   &  $\mathbf{0.886 \pm 0.005}$ & $\mathbf{0.687 \pm 0.012}$                         \\ \Xhline{2\arrayrulewidth}
\end{tabular}
\end{center}
\caption{The Signature-Wasserstein-1 metric and both the Extreme values metrics, IBEX35 dataset. The selected percentiles for the $\text{EV}_{+}$ and $\text{EV}_{-}$ were 95\% and 5\%, respectively.}
\end{table}
It is interesting to mention that, in terms of the Extreme values metric, the performance of the models considerably drops during the COVID-19 years. For example, for the Neural SDE the results of the $\text{EV}_{-}$ AUC in the period 2015-2019 is $0.758 \pm 0.033$, while in the period 2020-2022 is $0.633 \pm 0.004$.

\begin{table}[H]\setlength\belowcaptionskip{-10pt}
\begin{center}
\begin{tabular}{l c c c c}
\hline
             & $W_{1}$-a & $W_{1}$-b & $W_{1}$-c & $W_{1}$-d       \\ \Xhline{2\arrayrulewidth}
LSTM CWGAN    & $0.025 \pm 0.009$       & $0.021 \pm 0.004$   & $0.041 \pm 0.019$  &  $0.023 \pm 0.013$       \\
LSTM SigCWGAN & $0.014 \pm 0.002$         & $0.022 \pm 0.001$    & $\mathbf{0.027 \pm 0.004}$   & $0.025 \pm 0.001$           \\ 
NSDE SigCWGAN & $\mathbf{0.014 \pm 0.001}$ & $\mathbf{0.017 \pm 0.001}$   & $0.029 \pm 0.001$  & $\mathbf{0.016 \pm 0.002}$             \\ \Xhline{2\arrayrulewidth}
\end{tabular}
\end{center}
\caption{The four unordered Wasserstein-1 metrics, IBEX35 dataset.}
\end{table}

We can conclude that in this case the NSDE SigCWGAN clearly outperformed the rest of the models in terms of all metrics. 

\section{Conclusion}
In this paper, we have proposed the use of a Conditional Neural Stochastic Differential Equation as a conditional generator in the SigCWGAN algorithm, which offsets the great increase in terms of memory cost produced by the Montecarlo procedure needed for every sample in the minibatch.

We then tested in practice what was the gain and loss, in terms of computational time and memory cost, of first replacing the traditional WGAN algorithm with the SigCWGAN method, and then the traditional LSTM generator with a Neural SDE. We clearly showed that Neural SDEs trained with the SigCWGAN algorithm were the most balanced in terms of both resources cost. Finally, we compared their performance in four experiments with four different datasets, which allowed us to see that, in most cases, the Neural SDEs captured better some of the properties of the real datasets.

\section*{Acknowledgements}
The research of Josep Vives is partially supported by Spanish grant PID2020-118339GB-100 (2021-2024).

\section*{Declarations of Interest}

All authors report no conflicts of interest. The authors alone are responsible for the content and writing of the paper.

\medskip

\printbibliography

\newpage
\appendix
\section*{Supplementary Material}

\section{The signature transform}\label{signaturesection}
For completeness, we will give an elementary introduction to the \textit{signature transform}, and we will briefly see many of the properties that make it such a convenient transformation in machine learning. 

\subsection{Paths of bounded variation}

\begin{definition}[$p-$variation]\label{p variation not norm}
Let $p\geq 1$ be a real number, and $\norm{\cdot}$ any norm on $\mathbb{R}^{d}$. Let $x: [0,T] \to \mathbb{R}^{d}$ be a continuous path. The $p-$variation of $x$ is defined as
\begin{gather}
    \norm{x}_{p} = \Bigg[ \sup_{\mathcal{D}} \sum_{i=0}^{n-1} || x_{t_{i+1}}-x_{t_{i}} ||^{p} \Bigg]^{1/p}
\end{gather}
where the supremum is taken over the set of partitions of $[0,T]$, denoted as $\mathcal{D}$. The space of $d-$dimensional continuous paths of finite $p-$variation will be denoted as $\mathcal{V}^{p}([0,T];\mathbb{R}^{d})$.
\end{definition}
 We will equip $\mathcal{V}^{p}([0,T];\mathbb{R}^{d})$ with the following norm.
\begin{definition}[$p-$variation norm]
The $p-$variation norm of a path $x \in \mathcal{V}^{p}([0,T];\mathbb{R}^{d})$ is defined as 
\begin{gather}
    \norm{x}_{p-var} = \norm{x}_{p} + \sup_{t \in [0,T]} \norm{x_{t}}
\end{gather}
where $\norm{\cdot}$ is any norm in $\mathbb{R}^{d}$.
\end{definition}
For simplicity, we will only work with paths of finite $1-$variation, which are called of bounded variation. The reason for this is that in practice we will always be given a discrete stream of data, which can be converted into a continuous path by performing some interpolation scheme. The most common one is linear interpolation (with the resulting paths being of bounded variation), since then the signature is very easy to compute. This is what the Signatory package \parencite{kidger2021signatory} does, providing differentiable computations of the signature on the GPU.

Moreover, to define a signature of a general path of finite $p-$variation, we would need to introduce many new concepts of rough path theory, which is beyond the scope of this paper.

\subsection{The tensor algebra}

\begin{definition}[Tensor algebra of $\mathbb{R}^{d}$]
Let $(\mathbb{R}^{d})^{\otimes k}$ denote the $k$th tensor power of $\mathbb{R}^{d}$. By convention, $(\mathbb{R}^{d})^{\otimes 0} = \mathbb{R}$. We define the extended tensor algebra of $\mathbb{R}^{d}$ as
\begin{gather*}
     T((\mathbb{R}^{d})) = \{ \textbf{a} = (a_{0}, a_{1},...) \ | \ \forall k \geq 0, a_{k} \in (\mathbb{R}^{d})^{\otimes k} \}
\end{gather*}
We also define the truncated tensor algebra of order $N$ of $\mathbb{R}^{d}$, which is a linear subspace of $T((\mathbb{R}^{d}))$, as
\begin{gather*}
     T^{(N)}(\mathbb{R}^{d}) = \{ \textbf{a} = (a_{0}, a_{1},..., a_{N}) \ | \ \forall N \geq k \geq 0, a_{k} \in (\mathbb{R}^{d})^{\otimes k} \}
\end{gather*}
Note that $T^{(N)}(\mathbb{R}^{d})$ is a real vector space of dimension \begin{flalign}
\sum_{k=0}^{N} d^{k} = 
\begin{cases}
    N+1 & \text{if $d=1$} \\
    \frac{d^{N+1}-1}{d-1} & \text{if $d>1$} 
\end{cases}
\end{flalign}
\end{definition}
We will equip $T((\mathbb{R}^{d}))$ with an admissible norm, defined as follows.
\begin{definition}
We say that the extended tensor algebra $T((\mathbb{R}^{d}))$ is endowed with an admissible norm $\norm{\cdot}$ if the following conditions hold:
\begin{enumerate}
    \item For each $k \geq 1$, the symmetric group $\mathcal{S}_{k}$ acts by isometry on $(\mathbb{R}^{d})^{\otimes k}$, i.e.
    \begin{gather}
        \norm{\sigma v} = \norm{v}, \quad \forall v \in (\mathbb{R}^{d})^{\otimes k}, \ \forall \sigma \in \mathcal{S}_{k} 
    \end{gather}
    \item The tensor product has norm $1$, i.e. $\forall n, m \geq 1$,
    \begin{gather}
        \norm{v \otimes w} \leq \norm{v} \norm{w}, \quad \forall v \in (\mathbb{R}^{d})^{\otimes n}, \ w \in  (\mathbb{R}^{d})^{\otimes m}
    \end{gather}
\end{enumerate}
\end{definition}

\subsection{The signature of a path}
The signature transform provides a way of encoding any continuous path into an infinite sequence of statistics, which has many properties that make it a very desirable transformation in machine learning \parencite{https://doi.org/10.48550/arxiv.1603.03788}. 
\begin{definition}[Signature of a path]
Let $x = (x^{1},...,x^{d}) : [0,T] \to \mathbb{R}^{d}$ be a continuous path of bounded variation. The signature of $x$ is defined as the infinite collection of iterated integrals
\begin{gather*}
    sig(x) = \Bigg( \ \ \idotsint\limits_{0 < t_{1} < \dots < t_{k} < T} dx_{t_{1}} \otimes \dots \otimes dx_{t_{k}} \Bigg)_{k \geq 0} \in T((\mathbb{R}^{d}))
\end{gather*}
where the integration is in the Riemann-Stieltjes sense. By convention, the $k=0$ term is taken to be $1 \in \mathbb{R}$. For every fixed $k \geq 0$, the corresponding term is called the $k$th level of the signature.  
\end{definition}

One of the most important properties of the signature of a path is its uniqueness up to time-reparametrizations and translations. Specifically, we have the following result.
\begin{lemma}[\cite{Uniqueness_signature}]\label{the lemma}
Let $\mathcal{A}$ be a subset of $\mathcal{V}^{1}([0,T];\mathbb{R}^{d})$ such that all paths in $\mathcal{A}$ have the same initial value and have at least one monotone coordinate. Then the signature of a path completely determines it in $\mathcal{A}$.
\end{lemma}
In other words, after applying a couple of simple transformations to the original paths (namely the basepoint and time augmentations, see \cite{DBLP:journals/corr/abs-2006-00873}), the signature provides a perfect codification for any continuous path of bounded variation. We will denote such space as $\mathcal{A}^{1}([0,T];\mathbb{R}^{d})$.

In practice we are usually not able to compute all the terms of the signature, and are instead forced to calculate only a finite number of them.
\begin{definition}[Truncated signature of a path]
Let $x = (x^{1},...,x^{d}) : [0,T] \to \mathbb{R}^{d}$ be a continuous path of bounded variation. The truncated signature of order $N$ of $x$ is defined as the finite collection of iterated integrals
\begin{gather*}
    sig^{N}(x) = \Bigg( \ \ \idotsint\limits_{0 < t_{1} < \dots < t_{k} < T} dx_{t_{1}} \otimes \dots \otimes dx_{t_{k}} \Bigg)_{N \geq k \geq 0} \in T^{(N)}(\mathbb{R}^{d})
\end{gather*}
where $T^{(N)}(\mathbb{R}^{d})$ denotes the truncated tensor algebra of order $N$ of $\mathbb{R}^{d}$.
\end{definition}

It turns out that selecting the first levels of the signature is a good approximation, since its levels decay in size factorially. 

\begin{theorem}[Factorial decay, {\cite[Proposition 2.2]{Rough_paths}}]
Let $x:[0,T] \to \mathbb{R}^{d}$ be a continuous path of bounded variation. Let $sig_{k}(x)$ denote the $k$th level of the signature of $x$. Then, for any $k \in \mathbb{N}$ we have that
\begin{gather}\label{sup}
    \norm{sig_{k}(x)} \leq \frac{\norm{x}_{1-var}^{k}}{k!}
\end{gather}
\end{theorem}
A direct consequence is that one is able to approximate arbitrarily well the signature by the sequence of truncated signatures.
\begin{corollary}\label{app trunc}
Let $x:[0,T] \to \mathbb{R}^{d}$ be a continuous path of bounded variation. Then, for every $\epsilon>0$, there exists an $N \in \mathbb{N}$ such that 
\begin{gather}
    \norm{sig(x) - sig^{N}(x)} \leq \epsilon
\end{gather}
Moreover, if we restrict ourselves to a compact subset $K \subset \mathcal{V}^{1}([0,T];\mathbb{R}^{d})$, then the convergence is uniformly.
\end{corollary}

Another very important property of the signature is the universal non-linearity property, which states that the signature provides a basis for the space of continuous functions on path space. 

\begin{theorem}[Universal non-linearity, {\cite[Theorem 4.2]{https://doi.org/10.48550/arxiv.1809.09466}}]
Let $K \subset \mathcal{A}^{1}([0,T]; \mathbb{R}^{d})$ be a compact subset. Then
\begin{gather}
   \bigcup_{N \in \mathbb{N}} \big\{ x \mapsto \ell(sig^{N}(x)) \ | \ \ell: T^{(N)}(\mathbb{R}^{d+1}) \to \mathbb{R}^{v} \text{  is linear} \big\}
\end{gather}
is uniformly dense in the space of continuous maps from $K$ to $\mathbb{R}^{v}$, denoted by $\mathcal{C}(K;\mathbb{R}^{v})$.
\end{theorem}

\subsection{The signature of a stochastic process}

The last property that we will see states that the expected signature of a stochastic process that has a compact support completely determines its distribution. 

\begin{theorem}[\cite{Fawcett}]\label{expected sig}
    Let $\{X_{t}\}_{t \in [0,T]}$ and $\{Y_{t}\}_{t \in [0,T]}$ be two stochastic processes with compact support $K \subset \mathcal{A}^{1}([0,T]; \mathbb{R}^{d})$. Then, we have that
    \begin{enumerate}
        \item The expected signatures are well defined, $\mathbb{E}_{x\sim X}[sig(x)]$, $\mathbb{E}_{y \sim Y}[sig(y)] < \infty$.
        \item If $\mathbb{E}_{x\sim X}[sig(x)] = \mathbb{E}_{y \sim Y}[sig(y)]$, then $X$ and $Y$ are equal in the distribution sense. In other words, they have the same law.
    \end{enumerate}
\end{theorem}

\section{The Signature-Wasserstein-1 metric}\label{sigw1section}

We first recall the definition of the Wasserstein-1 distance between distributions defined on the same measurable space.   
\begin{definition}[Wasserstein-1 distance]
Let $\Pi(\mu, \nu)$ denote the set of all joint distributions on the measurable space $(\mathcal{X} \times \mathcal{X}, \mathcal{F} \otimes \mathcal{F})$ whose marginals are respectively $\mu$ and $\nu$. Then the Wasserstein-1 distance is defined as
\begin{gather}\label{wass}
    W_{1}(\mu, \nu) := \inf_{\gamma \in \Pi(\mu, \nu)} \mathbb{E}_{(x,y) \sim \gamma} [||x-y||_{1}]
\end{gather}
\end{definition}
In practice the infimum in (\ref{wass}) is highly intractable. Luckily for us, the Kantorovich-Rubinstein duality \parencite{optimal_transport} states that the Wasserstein-1 distance can be calculated as
\begin{gather}\label{supp}
    W_{1}(\mu, \nu) = \sup_{\norm{F}_{L} \leq 1} \Big\{ \mathbb{E}_{x \sim \mu}[ F(x)] - \mathbb{E}_{x \sim \nu}[ F(x)] \Big\}
\end{gather}
where the supremum is over all the $1-$Lipschitz functions $F: \mathcal{X} \to \mathbb{R}$. The most common approach is to approximate $F$ by a parameterized family of functions that are $1-$Lipschitz, like in the Wasserstein-GAN approach \parencite{pmlr-v70-arjovsky17a}. 

Now let $Z$ be a random variable taking values on a space $\mathcal{Z}$. Let $G_{\theta}: \mathcal{Z} \to \mathcal{X}$ be a map parameterized by $\theta$. Consider the pushforward conditional distribution in $\mathcal{X}$ that is given by $G_{\theta}(Z)$, which we denote as $\mathbb{P}_{\theta}$, and let $\mathbb{P}_{r}$ denote the distribution of the data. Then, our goal is to minimize an approximation of the Wasserstein-1 distance between $\mathbb{P}_{\theta}$ and $\mathbb{P}_{r}$.

In the Wasserstein GAN approach, the optimization problem that one tries to solve is
\begin{gather}\label{classical wgan}
    \min_{\theta \in \Theta} \max_{\psi \in \Psi} \Big\{ \mathbb{E}_{x \sim \mathbb{P}_{r}} [F_{\psi}(x)] - \mathbb{E}_{z \sim Z} [F_{\psi}(G_{\theta}(z))] \Big\}
\end{gather}
where $\{F_{\psi}\}_{\psi \in \Psi}$ is a family of $1-$Lipschitz functions. The training algorithm consists in performing $k$ steps on the parameters on the critic $F_{\psi}$, and once a good enough approximation of the Wasserstein-1 distance has been reached, perform one step on the parameters of the generator $G_{\theta}$. However, this method is very unstable and sensitive to the choice of the hyperparameters.
 
Remember that, in our case, we are interested in approximating distributions in path space, such that $\mathcal{X} = \mathcal{A}^{1}([0,T];\mathbb{R}^{d})$. Let $K \subseteq \mathcal{X}$ denote the union of the supports of $\mu$ and $\nu$. By definition of (\ref{supp}), there exists a sequence $F_{n}:K \to \mathbb{R}$ of functions that are $1-$Lipschitz such that they attain the supremum $W_{1}(\mu, \nu)$. 

If we assume that $K \subset \mathcal{X}$ is a compact subset, by the universal non-linearity property of the signature we have that, $\forall \epsilon>0$ and for each $F_{n}$ there exists a linear functional $\ell_{n}: T((\mathbb{R}^{d+1})) \to \mathbb{R}$ such that
\begin{gather}
    \sup_{x \in K} | F_{n}(x) - \ell_{n}(sig(x)) | \leq \epsilon
\end{gather}
This implies that the supremum in (\ref{supp}) can be attained by a sequence of linear functionals applied to the signature, and we have that (\ref{supp}) is equivalent to
\begin{flalign}\label{sigW}
    sigW_{1}(\mu, \nu) &= \sup_{\norm{\ell}_{L} \leq 1} \Big\{ \mathbb{E}_{x \sim \mu}[ \ell(sig(x))] - \mathbb{E}_{x \sim \nu}[ \ell(sig(x))] \Big\} \\
    &= \sup_{\norm{\ell}_{L} \leq 1} \ell \Big( \mathbb{E}_{x \sim \mu}[ sig(x)] - \mathbb{E}_{x \sim \nu}[ sig(x)] \Big)
\end{flalign}
where the expected signature is well defined, since we assumed that both measures have a compact support. Moreover, by using the truncated signatures and Corollary \ref{app trunc} we can approximate (\ref{sigW}) uniformly by 
\begin{gather}\label{signwgan}
    sig^{N}W_{1}(\mu, \nu) = \sup_{\norm{\ell}_{L} \leq 1} \ell \Big( \mathbb{E}_{x \sim \mu}[ sig^{N}(x)] - \mathbb{E}_{x \sim \nu}[ sig^{N}(x)] \Big)
\end{gather}
where now the linear functionals $\ell$ are defined in the truncated tensor algebra $T^{(N)}(\mathbb{R}^{d+1})$. Since the domain space is finite dimensional, we have that the Lipschitz constant of a linear functional $\ell: T^{(N)}(\mathbb{R}^{d+1}) \to \mathbb{R}$ is simply defined as the norm of its coefficients as an euclidean vector.  

It turns out that when we choose this to be the $L_{2}$ norm, we have that the optimization problem (\ref{signwgan}) admits a closed-form solution \parencite[Lemma A.3]{CSWGAN}, with the solution being
\begin{gather}\label{closedform}
    sig^{N}W_{1}(\mu, \nu) = \norm{ \mathbb{E}_{x \sim \mu}[ sig^{N}(x)] - \mathbb{E}_{x \sim \nu}[ sig^{N}(x)]}_{2}
\end{gather}
which is called the truncated Signature-Wasserstein-1 metric of order $N$. In the GAN setting, we can approximate the expected signatures of the model and the data by performing a Montecarlo simulation, i.e. by computing the empirical averages of the signatures. 

\section{The Conditional Signature-Wasserstein GAN approach}\label{cswgansection}
What if we have some knowledge we want to condition this distribution on? Let $X = \{X_{t}\}_{t \in [0,T_{x}]}$ and $Y = \{Y_{t}\}_{t \in [0,T_{y}]}$ be two stochastic processes taking values on 
\begin{gather}
    \mathcal{X} = \mathcal{A}^{1}([0,T_{x}];\mathbb{R}^{d_{x}}), \quad \quad \mathcal{Y} = \mathcal{A}^{1}([0,T_{y}];\mathbb{R}^{d_{y}})
\end{gather}
respectively. Let $p(X,Y)$ denote their joint distribution. Then, given any known trajectory $x \in \mathcal{X}$, we are interested in approximating the conditional distribution $p(Y | X=x)$, which we will compactly denote by $Y | x$.

In practice, we cannot approximate the expected signature of $Y | x$ by a Montecarlo procedure, since for every $x$ we are usually given only one sample. Therefore we must come up with a different method to estimate $\mathbb{E}_{y \sim Y|x}[sig(y)]$.

Just like in supervised learning\footnote{The difference with respect to supervised learning is that in this case we are interested in the conditional distribution, and not its expectation.}, we can formulate this problem as trying to approximate a map
\begin{gather}\label{complicated}
         \begin{array}{rccc}
    f \colon & \mathcal{X} & \to & \mathcal{P}(\mathcal{Y}) \\
    & x & \mapsto &  Y | x
\end{array}
\end{gather}
where $\mathcal{P}(\mathcal{Y})$ is the space of all stochastic processes taking values on $\mathcal{Y}$. If we restrict ourselves to family of stochastic processes with compact support, by Theorem \ref{expected sig} and Lemma \ref{the lemma} there will exist a unique function $\hat{f}: T((\mathbb{R}^{d_{x}+1})) \to T((\mathbb{R}^{d_{y}+1}))$
such that
\begin{gather}\label{hat_f}
    \hat{f}(sig(x)) = \mathbb{E}_{y \sim Y|x} [sig(y)]
\end{gather}
where we wrote $f(x)=Y|x$. If we assume that the domain of $f$ is a compact subset, we can use the uniform convergence of the truncated signatures to approximate (\ref{hat_f}) by 
\begin{gather}\label{here}
         \begin{array}{rccc}
    \hat{f}_{N,M} \colon & K \subset T^{(N)}(\mathbb{R}^{d_{x}+1}) & \to & T^{(M)}(\mathbb{R}^{d_{y}+1}) \\
    & sig^{N}(x) & \mapsto &  E_{y \sim  Y|x}[sig^{M}(y)]
\end{array}
\end{gather}
The image space of $\hat{f}_{N,M}$ is real and has finite dimension, so one can think of it as an euclidean space. Therefore we can use the universal non-linearity of the signature to approximate $\hat{f}_{N,M}$ by a linear functional $\ell_{N,M}$,
\begin{gather}\label{linear func}
         \begin{array}{rccc}
    \ell_{N,M} \colon & K \subset T^{(N)}(\mathbb{R}^{d_{x}+1}) & \to & T^{(M)}(\mathbb{R}^{d_{y}+1}) \\
    & sig^{N}(x) & \mapsto &  E_{y \sim  Y|x}[sig^{M}(y)]
\end{array}
\end{gather}

In conclusion, we have reduced a non-linear modelling problem between two continuous and infinite dimensional spaces in (\ref{complicated}) to a linear problem between two discrete and finite dimensional spaces. Moreover, the function $\ell_{N,M}$ represents a conditional expectation, and so we can use the classical linear regression framework to approximate it.

This is, given some data $\{x^{(i)},y^{(i)}\}_{i=1}^{n}$, we will assume that
\begin{gather}\label{linear regression}
    sig^{M}(y^{(i)}) = W \cdot sig^{N}(x^{(i)}) + \epsilon_{i}
\end{gather}
where 
\begin{itemize}
    \item $sig^{M}(y^{(i)})$ can be viewed as an $\frac{(d_{y}+1)^{M}-1}{d_{y}}$ dimensional real vector.
     \item $sig^{N}(x^{(i)})$ can be viewed as an $\frac{(d_{x}+1)^{N}-1}{d_{x}}$ dimensional real vector.
     \item $W$ is a real matrix of dimension $\frac{(d_{y}+1)^{M}-1}{d_{y}} \times \frac{(d_{x}+1)^{N}-1}{d_{x}}$.
     \item $\epsilon_{i} \sim \mathcal{N}(0, \sigma^{2}I)$ is i.i.d. Gaussian noise, with $I$ denoting the identity matrix.
\end{itemize}
and use any standard libraries to compute the weight matrix $W$. 

In practice, we will proceed as follows. Let $Z$ be a random variable taking values on a space $\mathcal{Z}$. Let $G_{\theta}: \mathcal{Z} \times \mathcal{X} \to \mathcal{Y}$ be a map parameterized by $\theta$. Then, for a known $x\in \mathcal{X}$, we will consider the pushforward conditional distribution in $\mathcal{Y}$ that is given by $G_{\theta}(Z, x)$. Just like in the unconditional case, the expected signature of $G_{\theta}(Z, x)$ is estimated through a Montecarlo procedure.

In order to estimate the expected signature of $Y|x$ from the set of given data, we simply perform the linear regression explained earlier, from which we obtain an estimate of (\ref{linear func}), that we denote by $\hat{\ell}$. Then we can obtain an approximation of $\mathbb{E}_{y \sim Y|x}[sig(y)]$ by 
\begin{gather}
    \hat{\mathbb{E}}_{y \sim Y|x}[sig^{M}(y)] = \hat{\ell}(sig^{N}(x))
\end{gather}

\section{Experimental Details}
The experimental details that were used in Section \ref{empirical analysis}, such as the model architectures and the selected hyperparameters, can be found in the following pages.

\subsection{Architectures for the resources cost analysis}\label{details resources}

\textbf{LSTM Conditional WGAN} The LSTMs of the conditional generator had 5 hidden layers of width 32, with 5 dimensional random noise, and had 77,089 learnable parameters. The LSTMs of the critic also had 5 hidden layers of width 32, with 76,609 learnable parameters.
\vspace{0.3cm}

\textbf{LSTM Conditional Sig-WGAN} The architecture of the generator was the same as the one we just described for the LSTM Conditional WGAN generator. The signature transform was truncated at depth 5 and 4, for the input and output paths, respectively. We also applied the cumulative sum transformation\footnote{Although the truncated signatures approximate arbitrarily well the signature transform, this does not mean that some crucial information might be lost by not computing higher levels. In order to mitigate this, usually some transformations are applied to the given stream, see \cite{DBLP:journals/corr/abs-2006-00873}.}. Overall, the dimension of both truncated signatures was 363 and 120, respectively. 
\vspace{0.3cm}

\textbf{Neural SDE Conditional Sig-WGAN} The number of hyperparameters in a conditional Neural SDE is slightly larger. Since it is a little bit hard to keep track of all of them, we will mention them along the corresponding notation we used in Definition \ref{cnsde}.

The size $d_{z}$ of the Neural SDE was 92. The initial random noise size $d_{v}$ was 16 and the network $\xi_{\theta}^{2}$ was formed by one hidden layer of size 32. The network $\xi_{\theta}^{1}$ was simply a linear function.
The initial condition of the SDE was formed by a random part $k$ of size 8 and a deterministic part $d_{z}-k$ of size 84. The diffusion $g_{\theta}$ was chosen to be of general type. The dimension of the Wiener process $d_{w}$ was 10. Both the drift $f_{\theta}$ and diffusion $g_{\theta}$ had a hidden layer of width 32. The only activation function we used was the hyperbolic tangent, tanh. The total number of learnable parameters was 79,273.

For the Neural SDE the signatures in the CSig-WGAN algorithm were of the same order and with the same transformations as the LSTM Conditional Sig-WGAN model. The transformation $\phi(x)$ of the input paths $x$ was set to be the signature transform, with the same truncation order and transformation the ones used in the CSig-WGAN algorithm.

\subsection{Architectures and hyperparameters for the experiments}
In this section we will detail the architectures and hyperparameters that were used for each experiment in Section \ref{experiments}, which were selected by performing an informal grid search.

First we will list the ones that were common to all the datasets.

\subsubsection{Common hyperparameters}\label{common hyper}
The optimizers that were used were as follows: for the LSTM CWGAN, following \cite{pmlr-v70-arjovsky17a} we used RMSprop. For the models trained with the CSig-WGAN algorithm, we used Adam. The learning rate was always set to $10^{-3}$. 

In all cases the signature transform was truncated at depth 5 and 4, for the input and output paths, respectively. We also applied the cumulative sum transformation, and we normalized each dimension so that it had zero mean and unit variance. Overall, the dimension of both truncated signatures was 363 and 120, respectively. The transformation $\phi(x)$ of the Neural SDEs was set to be the signature transform, with the same truncation order and transformation the ones used in the CSig-WGAN algorithm.

\subsubsection{AR(p) dataset}\label{details arp}
The training set was formed by 14,900 pairs of time series data $(x,y)$. The validation set was formed by 2,400 samples. The test set was formed by 12,400 samples. We normalized the data with the mean and standard deviation of the input streams in the training set.

The architectures and other hyperparameters for each model were as follows.

\textbf{LSTM WGAN} The LSTMs in the generator had 5 layers with hidden size equal to 32, with 5 dimensional random noise. The LSTMs in the critic had 4 layers with hidden size 32. The number of parameters of the generator was 77,089, while the number of parameters of the critic was 59,713. 

\textbf{LSTM Sig-WGAN} As we already explained, the generator was the same as in the LSTM WGAN model.

\textbf{Neural SDE Sig-WGAN } The size $d_{z}$ of the Neural SDE was 48. The initial random noise size $d_{v}$ was 16 and the network $\xi_{\theta}^{2}$ was formed by one hidden layer of size 32. The network $\xi_{\theta}^{1}$ was simply a linear function.

The initial condition of the SDE was formed by a random part $k$ of size 16 and a deterministic part $d_{z}-k$ of size 32. The diffusion $g_{\theta}$ was chosen to be of diagonal type, and therefore the dimension of the Wiener process $d_{w}$ was the same as the size of the solution, 48. Both the drift $f_{\theta}$ and diffusion $g_{\theta}$ had one hidden layer of width 84. The only activation function we used was the hyperbolic tangent, tanh. Following \cite{kidger2021neuralsde}, we also applied a final tanh to the vector fields. The total number of learnable parameters was 29,161.  

For the LSTM WGAN and the Neural SDE Sig-WGAN models, the batch size was set to 528. Due to memory constrains, for the LSTM Sig-WGAN the batch size was set to 228.

\subsubsection{Seattle Weather dataset}\label{details sw}
The train set was formed by 15,122 pairs of time series data $(x,y)$. The validation set was formed by 3,781 samples. The test set was formed by 6,468 samples. The test set was extracted from a different time period than the train and validation sets (out-of-time samples). We normalized the data with the mean and standard deviation of the input streams in the training set.

The architectures and other hyperparameters for each model were as follows.

\textbf{LSTM WGAN} The LSTMs in the generator had 5 layers with hidden size equal to 32, with 5 dimensional random noise. The LSTMs in the critic had 4 layers with hidden size 36. The number of parameters of the generator was 77,089, while the number of parameters of the critic was 75,241. 

\textbf{LSTM Sig-WGAN} The generator was the same as in the LSTM WGAN model.

\textbf{Neural SDE Sig-WGAN } The architecture was the same as the one we detailed in Section \ref{details arp}, with the exception that we set the hidden size of the SDE to $d_{z} = 64$. Just like in the previous experiment we applied a final tanh to the vector fields of the SDE. The total number of parameters was 35,113.

For the Neural SDE Sig-WGAN models, the batch size was set to 724. For the LSTM WGAN, it was set to 528.  Due to memory constrains, for the LSTM Sig-WGAN the batch size was set to 228.
\subsubsection{Forex dataset}\label{details forex}
The train set was formed by 15,000 pairs of time series data $(x,y)$. The validation set was formed by 5,000 samples. The test set was formed by 10,000 samples. The test set was extracted from a different time period than the train and validation sets (out-of-time samples). As it is usually done in this kind of datasets, we applied the logarithm transformation to the data. After that, we normalized the data with the mean and standard deviation of the input streams in the training set.

The architectures and other hyperparameters for each model were as follows.

\textbf{LSTM WGAN} The architecture of the generator was the same as the one we detailed in 
the Seattle Weather dataset experiment. However, in this case the architecture of both LSTMs of the critic was set to hidden size equal to 32 and a number of layers equal to 5. The number of parameters of the critic was 234,113.

\textbf{LSTM Sig-WGAN} The generator was the same as in the LSTM WGAN model.

\textbf{Neural SDE Sig-WGAN } The architecture was the same as the one we detailed in 
the Seattle Weather dataset experiment. 

For the LSTM WGAN and the Neural SDE Sig-WGAN, the batch size was set to 528. Due to memory constrains, for the LSTM Sig-WGAN the batch size was set to 128.

\subsubsection{IBEX35 dataset}\label{details ibex}
The train set was formed by 4,296 pairs of time series data $(x,y)$. The validation set was formed by 1,074 samples. The test set was formed by 1,944 samples. The test set was extracted from a different time period than the train and validation sets (out-of-time samples). We normalized the data with the mean and standard deviation of the input streams in the training set.

The architectures and other hyperparameters for each model were as follows.

\textbf{LSTM WGAN} The LSTMs in the generator had 2 layers with hidden size equal to 64, with 5 dimensional random noise. The LSTMs in the critic had 4 layers with hidden size 32. The number of parameters of the generator was 101,953, while the number of parameters of the critic was 59,713. 

\textbf{LSTM Sig-WGAN} The generator was the same as in the LSTM WGAN model.

\textbf{Neural SDE Sig-WGAN } The size $d_{z}$ of the Neural SDE was 120. The initial random noise size $d_{v}$ was 16 and the network $\xi_{\theta}^{2}$ was formed by one hidden layer of size 48. The network $\xi_{\theta}^{1}$ was simply a linear function.

The initial condition of the SDE was formed by a random part $k$ of size 56 and a deterministic part $d_{z}-k$ of size 64. The diffusion $g_{\theta}$ was chosen to be of diagonal type, and therefore the dimension of the Wiener process $d_{w}$ was the same as the size of the solution, 64. Both the drift $f_{\theta}$ and diffusion $g_{\theta}$ had one hidden layer of width 96. The only activation function we used was the hyperbolic tangent, tanh. Just like in the previous experiments we applied a final tanh to the vector fields of the SDE. The total number of learnable parameters was 73,489.  

For the LSTM WGAN and the Neural SDE Sig-WGAN, the batch size was set to 1024. Due to memory constrains, for the LSTM Sig-WGAN the batch size was set to 528.

\section{Computing infrastructure}

All experiments were run on a computer that had Windows 11 Home as the operative system, equipped with an AMD Ryzen 7 5800X 8-Core Processor, 16GB of RAM and an Nvidia GeForce RTX 3060 with 12GB of memory.

The main python libraries that were used are listed below:
\begin{itemize}
    \item Pytorch 1.9.0+cu111 as the main deep learning framework \parencite{NEURIPS2019_9015}.
    \item Signatory 1.2.6, which provides differentiable computations of the signature on the GPU \parencite{kidger2021signatory}.
    \item torchsde 0.2.5, which provides stochastic differential equation (SDE) solvers with GPU support and efficient backpropagation \parencite{torchsde}.
\end{itemize}
We highlight that, in order to use the Signatory package, one needs to have a Pytorch version that is no older than 1.9.0.

\end{document}